%% file: root.tex
\documentclass[letterpaper, 10 pt, journal, twoside]{ieeetran}  

\IEEEoverridecommandlockouts                              

\let\labelindent\relax

\usepackage[
    style=ieee,
    doi=false,
    isbn=false,
    url=false,
    eprint=false,
    backend=bibtex,
    natbib=true,
    minnames=1,
    maxcitenames=1
    ]{biblatex}

\makeatletter
\def\ps@IEEEtitlepagestyle{%
  \def\@oddfoot{\mycopyrightnotice}%
  \def\@oddhead{\hbox{}\@IEEEheaderstyle\leftmark\hfil\thepage}\relax
  \def\@evenhead{\@IEEEheaderstyle\thepage\hfil\leftmark\hbox{}}\relax
  \def\@evenfoot{}%
}
\def\mycopyrightnotice{%
  \begin{minipage}{\textwidth}
  \centering \scriptsize
  Copyright~\copyright~2022 IEEE. Personal use of this material is permitted. Permission from IEEE must be obtained for all other uses, in any current or future media, including\\reprinting/republishing this material for advertising or promotional purposes, creating new collective works, for resale or redistribution to servers or lists, or reuse of any copyrighted component of this work in other works by sending a request to pubs-permissions@ieee.org.
  \end{minipage}
}
\makeatother

\usepackage[shortcuts]{extdash}
\usepackage{graphicx}
\usepackage{mathtools}
\usepackage{dcolumn}
\usepackage{gensymb}
\usepackage{perl_acronyms}
\usepackage{xcolor}
\usepackage{subcaption}
\usepackage{multirow}
\usepackage{amsfonts}
\usepackage{booktabs}
\usepackage{siunitx}
\usepackage[inline]{enumitem}
\usepackage{array}
\usepackage{flushend}
\newcolumntype{P}[1]{>{\centering\arraybackslash}p{#1}}

\newcommand{\revision}[1]{\textcolor{black}{#1}}

\author{Gideon Billings$^{1}$, Richard Camilli$^{2}$, Matthew Johnson-Roberson$^{3}$%
\thanks{
This work was supported by NASA grant NNX16AL08G and by the National Science Foundation  under  grants  IIS-1830660  and  IIS-1830500.}
\thanks{$^{1}$Gideon Billings is with Department of Naval Architecture and Marine Engineering, University of Michigan, 2600 Draper Dr. Ann Arbor, MI 48109, USA
        {\tt\footnotesize gidobot@umich.edu}}%
\thanks{$^{2}$Richard Camilli is with Applied Ocean Physics and Engineering, Woods Hole Oceanographic Institution, Deep Submergence Laboratory Woods Hole, MA 02543, USA}%
\thanks{$^{3}$Matthew Johnson-Roberson is with The Robotics Institute, Carnegie Mellon University, 5000 Forbes Avenue Pittsburgh, PA 15213-3890, USA}%
\thanks{$^{4}$https://github.com/gidobot/UWslam\_dataset}
\thanks{Digital Object Identifier (DOI): 10.1109/LRA.2022.3176448.}
}

\title{Hybrid Visual SLAM for Underwater Vehicle Manipulator Systems}

\begin{document}

\maketitle

\begin{abstract}
\label{sec:abstract}
\input{text/abstract.tex}
\end{abstract}


\section{Introduction}
\label{sec:introduction}
\input{text/introduction.tex}
\section{Related Work}
\label{sec:related}
\input{text/relatedwork.tex}

\section{Method}
\label{sec:method}
\input{text/method.tex}

\section{Evaluation}
\label{sec:results}
\input{text/results.tex}

\section{Conclusion}
\label{sec:conclusion}
\input{text/conclusion.tex}


\renewcommand{\bibfont}{\normalfont\small}
\printbibliography

\end{document}

%% file: text/abstract.tex
This paper presents a novel visual \revision{feature based} scene mapping method for underwater vehicle manipulator systems (UVMSs), with specific emphasis on robust mapping in natural seafloor environments. Our method uses GPU accelerated SIFT features in a graph optimization framework to build a feature map. The map scale is constrained by features from a vehicle mounted stereo camera, and we exploit the dynamic positioning capability of the manipulator system by fusing features from a wrist mounted fisheye camera into the map to extend it beyond the limited viewpoint of the vehicle mounted cameras. Our hybrid SLAM method is evaluated on challenging image sequences collected with a UVMS in natural deep seafloor environments of the Costa Rican continental shelf margin, and we also evaluate the stereo only mode on a shallow reef survey dataset. Results on these datasets demonstrate the high accuracy of our system and suitability for operating in diverse and natural seafloor environments. \revision{We also contribute these datasets$^3$ for public use.}

%% file: text/introduction.tex

Exploration vehicles for remote environments, such as rovers, planetary landers, or underwater Remotely Operated Vehicles (ROVs) are often equipped with manipulator systems for collecting samples, placing sensors, or otherwise interacting with the environment. These systems largely rely on direct tele-operation or manually scripted commands to execute manipulation tasks, due to the risks associated with acting in unstructured and often complex remote environments. Despite these risks, there are some remote environments, such as Europa, the icy moon of Jupiter, so distant that any kind of tele-operation or pre-scripted manipulator control is highly impractical. Considering environments closer to home, the deep ocean is a domain of intensive scientific research and exploration, but operation of deep submergence ROVs with attendant support ships and pilots is costly and availability is limited. \revision{Also, human error during tele-operated control can lead to damage to the vehicle or loss of valuable mission time.} These considerations motivate the automation of manipulator systems for exploration vehicles, to enable complex workspace interactions in communication limited or denied environments \revision{and to support tele-operated procedures}. This advancement can reduce the overhead requirements of supporting tele-operation infrastructure (including associated human labor and hotel functions), thereby increasing the availability of robotic systems for scientific research. Critical to achieving safe and robust autonomy of such vehicle-manipulator systems is scene perception and reconstruction. In this work, we address the problem of feature based 3D scene mapping for underwater vehicle-manipulator systems (UVMSs). A key innovation of our mapping system is the fusion of feature points from both a vehicle mounted stereo camera and a dynamically positioned manipulator mounted fisheye camera into the same mapping framework. In situations where a UVMS's movement is limited or risky, our method addresses the challenge of limited viewpoints from the vehicle mounted cameras and incomplete scene reconstruction due to shadowing from scene structure by enabling the wrist mounted camera to dynamically extend the map beyond the vehicle fixed camera views and fill in shadowed areas of the scene.

This work makes the following contributions:
\begin{enumerate}
    \item To our knowledge, the first SLAM system, designed for manipulator systems, that fuses a manipulator mounted fisheye camera into the same map with a vehicle mounted stereo camera.
    \item An adaptation of the ORB-SLAM2 framework to GPU accelerated SIFT features, with improved odometer based tracking and real-time performance.
    \item An evaluation of our method on both shallow reef and natural deep seafloor environments, where our method achieves good performance and standard ORB-SLAM2 fails. The evaluation datasets are also published with this paper$^4$.
\end{enumerate}

%% file: text/relatedwork.tex
3D scene mapping is a very mature problem in computer vision and robotics, and a rich body of literature has been generated from decades of study on the topic. Here we present a review of the works which we consider most relevant to our developed method and from which we took inspiration in our approach.

\subsection{Feature Based Visual SLAM}

Since its inception, ORB-SLAM2~\cite{mur2017orb}, remains one of the most widely adopted and complete feature based SLAM systems, demonstrating that a bundle adjustment approach can attain more accurate camera localization than direct methods or ICP, with the advantage of being less computationally expensive. Given the established robustness of ORB-SLAM2 across a variety of applications and camera systems, the efficient computational performance based on a parallel thread architecture, and the demonstrated accuracy of keyframe based bundle adjustment for pose estimation, we chose to develop our method based on this framework.

CoSLAM~\cite{zou2012coslam} proposed an innovative solution for fusing multiple synchronized but independently moving monocular cameras into a single framework that can also differentiate between dynamic and static feature points. We took inspiration from this approach in our method design, with the key differences being our use of stereo features to constrain the map scale, our fusion of independent hybrid camera frames into the same map (i.e. the manipulator mounted fisheye camera and a vehicle mounted perspective stereo camera), and the specific adaptations of our method to underwater environments.

\subsection{Underwater SLAM}

Significant progress has been made in underwater vision applied to large scale survey reconstructions~\cite{Johnson-Roberson:2016aa}, terrain aided navigation~\cite{negre_cluster-based_2016, ferrera_real-time_2020, hidalgo2018monocular}, and ship hull inspection~\cite{ozog2017mapping}. However, dense scene reconstruction methods generally process the image data offline, and methods designed for navigation generally provide very sparse feature maps if any. In contrast, our method emphasises real-time scene mapping, suitable for natural seafloor environments, that is robust to underwater visual effects and provides an optimized feature map and camera pose graph that can underlie dense reconstruction methods.

\citet{negre_cluster-based_2016} proposed a stereo based SLAM method specifically designed for operating in underwater feature-poor environments. The map is constructed as a pose graph connecting to feature clusters. For inter-frame pose estimation of non keyframes, they used the VISO2 stereo odometer~\cite{Geiger2011IV}, which they found to perform better than the tracking stage in ORB-SLAM. For detecting loop closures, this method generated a HALOC~\cite{carrasco2016global} signature for each feature cluster, which can be efficiently matched across very large image sets and does not require a prior training step like a bag of words representation. This work informed our choice of using a modified version of VISO2 for the initial inter-frame pose estimations. While their method was tailored specifically to the problem of localization through the optimization of keypoint cluster locations, our method, based on ORB-SLAM2, optimizes the location of the individual map points, which is desirable for scene reconstruction. \citet{hidalgo2018monocular} studied off the shelf monocular ORB-SLAM applied in different shallow oceanic underwater environments. Their results showed that ORB-SLAM performed well in structured or feature rich environments with adequate lighting and low flickering. However, ORB-SLAM performed poorly in areas with highly dynamic lighting, large numbers of moving objects, or low textured regions such as sand beds.

\subsection{Kinematics in SLAM}

Prior research has utilized eye-in-hand based SLAM, where a camera is mounted near the end effector of a manipulator. ARM-SLAM~\cite{klingensmith2016articulated} used a Kinect depth sensor mounted on a manipulator with a fixed base to capture point clouds of the scene and fuse them into a reconstruction using a method based on Kinect Fusion. SKCLAM~\cite{li2019slam} used feature based pose tracking with an RGB-D camera on the endeffector to calibrate the full kinematic parameters of an industrial manipulator with a fixed base. Point clouds from the RGB-D camera were integrated to construct a 3D map. \citet{chen20213d} used ORB-SLAM3 and a stereo camera on a mobile manipulator to map an orchard. Unlike these prior works, our method fuses features from both an independent manipulator mounted fisheye camera and a vehicle mounted stereo in a common feature graph. We use a monocular camera on the wrist rather than relying on a depth sensor, which would be very bulky to fit in a pressure rated housing for mounting on the manipulator. \citet{7759682} proposed a method for calibrating a dynamic camera cluster, where one camera is articulated with respect to the other cameras in the system. They demonstrated multi-camera SLAM with one camera mounted on a pan-tilt unit, assuming accurate calibration of the pan-tilt unit's extrinsics. In contrast, our method is demonstrated with the manipulator camera having 5-DoF actuation using very high baseline to the vehicle camera, and without relying on accurate extrinsic measurement of the articulated camera. To our knowledge, this is the first successful demonstration of an eye-in-hand SLAM method on mobile underwater manipulator platforms in natural deep ocean environments.

%% file: text/method.tex
In this section, we highlight the innovations made to adapt the ORB-SLAM2 system to SIFT features, the underwater environment, and the hybrid imagery. For details on the system architecture that remain unchanged from ORB-SLAM2, we defer the reader to~\cite{mur2017orb}.

\begin{figure}
    \centering
    \includegraphics[width=1.0\linewidth]{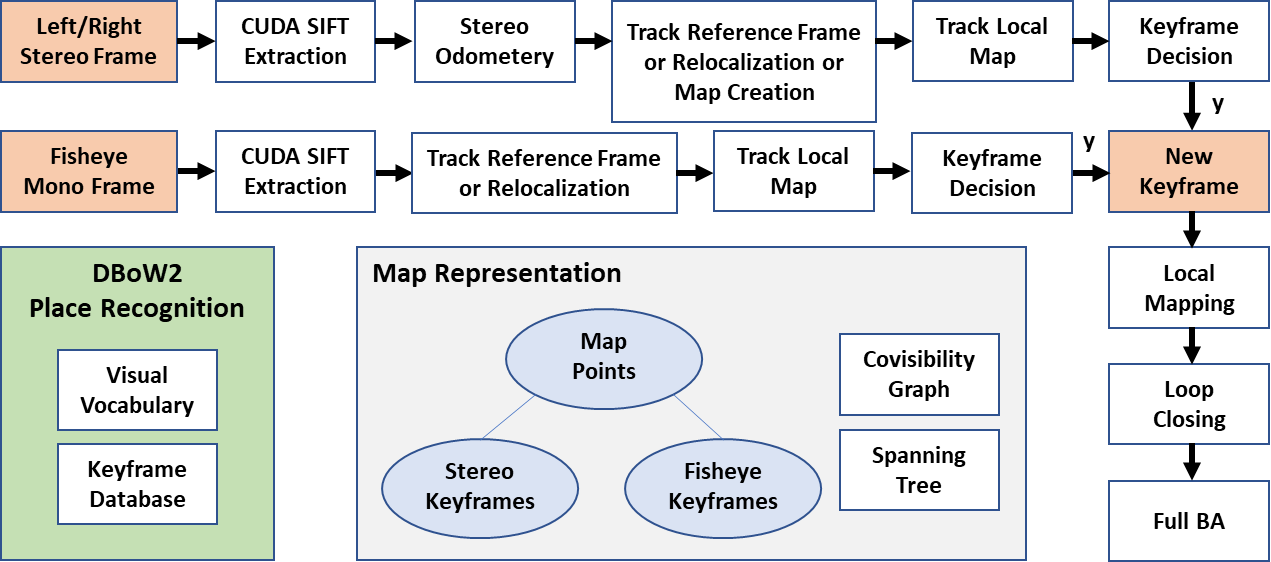}
    \caption{System block diagram}
    \vspace{-10pt}
    \label{fig:diagram}
\end{figure}

Figure~\ref{fig:diagram} shows a high level block diagram of the hybrid SLAM system, where our method retains the same four threaded architecture as the original ORB-SLAM2. Crucial modifications were made in the tracking thread, which follows the top horizontal flow of the diagram, with separate functional flow branches for stereo and monocular fisheye frames. Both stereo and fisheye frames share a common keyframe representation which is processed through the local mapping, loop closing, and full bundle adjustment threads. The core of our system is the feature based stereo mapping framework, which can be operated stand-alone or in a hybrid mode, where frames from an independently moving fisheye camera are fused into the same map.

The constructed map is represented as a covisibilty graph of optimized keyframe and keypoint poses, with factors between keyframes formed through common keypoint observations. Like ORB-SLAM2, the covisibility graph is used to retrieve a local neighborhood of keypoints for the tracking and local mapping stages and forms the graph structure for the bundle adjustment optimizations. A minimum spanning tree is also maintained, which connects every keyframe to the neighbor with the maximum number of shared keypoint observations. The spanning tree is used to propagate keyframe pose optimizations from full bundle adjustment to new keyframes that were not included during the optimization. A DBoW2 module~\cite{GalvezTRO12}, \revision{that we} adapted to SIFT features, is used for place recognition during relocalization and loop closing.

\subsection{Hybrid Camera System}

The hybrid camera system is specifically tailored to mobile manipulator systems, with a stereo camera mounted on the vehicle and an independent but synchronized fisheye camera mounted on the manipulator wrist. In our evaluations, the stereo pair uses a pinhole camera model, \revision{calibrated with the ROS~\cite{quigley2009ros} camera calibration package}, and the fisheye camera uses the Kannala-Brandt~\cite{kannala2006generic} model, \revision{calibrated with the Kalibr toolbox~\cite{kalibr2006}. Underwater checkerboard imagery was collected for both cameras to perform the calibrations. We adapt the camera model code from ORB-SLAM3~\cite{campos2021orb} to support our hybrid camera system. In our notation, we represent the camera projection function as $\pi_{(\cdot)}$, where $\pi_{m}$ and $\pi_{s}$ are the left monocular stereo camera and rectified stereo projection functions respectively as defined in~\cite{mur2017orb} and $\pi_{f}$ is the calibrated fisheye projection function from~\cite{kannala2006generic}.}

\subsection{Feature Representation}

While ORB-SLAM2 uses ORB~\cite{rublee2011orb} features, ORB performs poorly in many underwater environments compared to other feature types. We conducted an analysis of the matching performance of different feature types in the underwater domain, presented in the results section, which motivated our choice of the SIFT~\cite{lowe2004distinctive} feature for our system. We adopted CudaSIFT~\cite{bjorkman2014detecting}, which is currently one of the fastest GPU accelerated SIFT implementations, for real-time feature extraction.

\subsection{System Initialization}

On system startup, the first keyframe \revision{$K_0$} is created from the first stereo frame \revision{$F_{s,0}$} that retains at least 8\% of the maximum number of features \revision{$N$ extracted from the images}. \revision{$K_0$} is set as the origin of the initial map, which is constructed from all of the stereo keypoints. After the map is initialized, new keyframes are added from both the stereo and monocular fisheye frames, with the map scale constrained by the initial and new stereo map points.

\subsection{Stereo Odometery}

Similar to~\cite{negre_cluster-based_2016}, we found that the tracking stage of ORB-SLAM2 failed on our underwater datasets, even when adapted to SIFT features. A considerable limitation of the ORB-SLAM2 tracking stage is a constant velocity model, which has poor accuracy at the low frame rates typical for underwater imaging systems. ~\citet{negre_cluster-based_2016} used the VISO2 stereo odometer for initial frame pose estimation. We took inspiration from this and also adopted VISO2 for our system. However, we found that off-the-shelf VISO2 failed to track our underwater stereo dataset, due to poor performance of the simple blob and corner response features, described in the Sobel operator space. We modified VISO2 to use CudaSIFT features, which are extracted once for each image and then propagated through the rest of the SLAM pipeline for efficient computation. While the original VISO2 implementation used a search window to circularly match features across the current and previous stereo pair, we use GPU accelerated brute force matching, followed by circular filtering for improved computational performance. In this scheme, brute force matching is applied between the \textit{previous left} and \textit{previous right} frames, \textit{previous right} and \textit{current right} frames, \textit{current right} and \textit{current left} frames, and \textit{current left} and \textit{previous left} frames. A feature is accepted only if the same feature is matched across all image pairs in a circular fashion. Like in the original VISO2 implementation, feature matches between a  \textit{left} and \textit{right} stereo image pair are further filtered by an epipolar constraint of 1 pixel error tolerance. However, we found the outlier removal step of the original VISO2 by 2d Delaunay triangulation to be too restrictive in high rugosity coral reef imagery, resulting in the filtering of many correct feature correspondences. Through extensive experimentation, we found the circular matching and epipolar constrained filtering steps were sufficient for removing the majority of outliers before processing the matches through the ego-motion estimation stage. \revision{The output transform $\widetilde{T}_{s,i}$ of the odometer is the estimated ego-motion of the current stereo frame $F_{s,i}$ with respect to the previous stereo frame $F_{s,i-1}$.}

\subsection{Tracking}
Given an initialized system, the current \textbf{stereo frame} $F_{s,i}$ is processed through the tracking stage in the following steps:
\begin{enumerate}
\item Process $F_{s,i}$ through the odometer to extract CudaSIFT features and estimate the ego-motion $\widetilde{T}_{s,i}$ with respect to the previous stereo frame.
\item Initialize pose of $F_{s,i}$ as $T_{s,i} = \widetilde{T}_{s,i}T_{s,i-1}$, where $T_{s,i} \in SE(3)$.
\item Project the map points tracked in previous frame into the current frame using projection function $x=\pi_{m}(T_{s,i}X)$, where $X$ is the vector coordinate of the map point in the world reference frame, and search feature matches within window regions around the projected points in a brute force manner.
\item Formulate the iterative pose optimization:
\begin{equation}
\underset{T_{s,i}}{argmin}\sum_{j \in \chi}\rho(\left \| x_{(\cdot)}^j - \pi_{(\cdot)}(T_{s,i}X^j) \right \|_{\Sigma }^{2})
\end{equation}
where $j \in \chi$ is the set of all map point matches, $\rho$ is the robust Huber cost function, $\Sigma$ is the covariance matrix associated with the map point scale, and keypoints can be monocular with $x_{m}^j \in R^2$ for projection $\pi_{m}$ or stereo with $x_{s}^j \in R^3$ for projection $\pi_{s}$. Only retain inlier map point matches following the optimization.
\item If steps 3-4 fail to find enough map point correspondences, track $F_{s,i}$ to the map points tracked in the reference stereo keyframe $K_s$ using BoW vocabulary levels to guide matching followed by pose optimization as in step 4.
\item If few map point matches are found in steps 3-5, set $T_{s,i}$ to the odometery estimated pose of step 2.
\item Search additional map point matches in a local window of keyframes that share observations of the current map point matches and may include both stereo and fisheye frames, and repeat the pose optimization of step 4.
\end{enumerate}

If the stereo tracking stage fails completely, the system enters relocalization mode until tracking is recovered for a stereo frame, as in~\cite{mur2017orb}.

In hybrid mode, the current monocular \textbf{fisheye frame} $F_{f,i}$ is processed through the tracking stage in the following steps, but only after the current stereo frame is successfully tracked:
\begin{enumerate}
\item Track $F_{f,i}$ to the map points tracked in the reference fisheye keyframe $K_f$ using BoW vocabulary levels to guide matching followed by pose optimization as in step 4 of the stereo tracking. The optimization formulation is the same as Eq. (1), with fisheye projection $\pi_f$ and $x_{f}^j \in R^2$.
\item If step 1 fails, $F_{f,i}$ is tracked to the map points in the reference stereo keyframe $K_s$.
\item Search additional map point matches in a local window of keyframes, that may include both stereo and fisheye frames, and repeat the pose optimization step.
\end{enumerate}

If tracking fails for the fisheye frame but not the current stereo frame, the system enters relocalization mode for only the fisheye camera, while continuing mapping of the stereo frames. In this relocalization mode, the current fisheye frame is first attempted to be matched against all keyframes in the map using the BoW place recognition to identify match candidates. If place recognition fails, tracking of the current fisheye frame is then attempted against the current stereo reference keyframe, proceeding from step 2 above.
During the local mapping stage for both stereo and fisheye frames, the reference keyframe for each is updated to the keyframe that shares the most feature matches, agnostic to the type of keyframe (i.e. stereo or fisheye). When a new keyframe is inserted, it is made the reference keyframe for the next frame of the same type.

During relocalization or when the fisheye frame is tracked against the reference stereo frame, a perspective-n-point (PnP) solver is constructed to estimate an initial pose. Like \cite{campos2021orb}, we adopt the Maximum Likelihood Perspective-n-Point algorithm (MLPnP)~\cite{urban2016mlpnp}, which uses projective rays in the optimization that are agnostic to the camera model, in order to accurately optimize the feature correspondences between the hybrid fisheye and perspective stereo frames.

\subsection{Inserting New Keyframes}

New stereo and fisheye keyframes are decided following the same scheme as \cite{mur2017orb} for stereo and monocular keyframes respectively, with some thresholds tuned for lower framerates and higher keypoint counts. 
When a new keyframe is inserted, new map points are triangulated and added into the map. For each of these keypoints the maximum and minimum distances that the point can be detected in a frame are calculated based on the scale of the keypoint in the reference keyframe. With the hybrid camera system, the scale of the keypoint can be different at the same distance, depending on which type of frame observes the keypoint. We resolve this ambiguity by normalizing the keypoint scale factor by the focal length of the observing frame. This normalization enables consistent keypoint scale prediction and comparison between hybrid frames.
\revision{
As in \cite{mur2017orb}, local bundle adjustment is performed on a set of covisible keyframes $\mathcal{K}_L$ and all points seen in those keyframes $\mathcal{P}_L$ after a new keyframe is inserted into the map, with the optimization formulated as follows:
\begin{equation}
\underset{X^i,T_{l}}{argmin}\sum_{k \in \mathcal{K}_L\cup \mathcal{K}_F}\sum_{j \in \chi_k}\rho(\left \| x_{(\cdot)}^j - \pi_{(\cdot)}(T_kX^j) \right \|_{\Sigma }^{2})
\end{equation}
where $i \in \mathcal{P}_L$, $l \in \mathcal{K}_L$, $\chi_k$ is the set of point matches between $\mathcal{P}_L$ and keypoints in keyframe $k$, and $\mathcal{K}_F$ is all other keyframes observing $\mathcal{P}_L$ but not in $\mathcal{K}_L$ which contribute to the cost but are fixed in the optimization. Keyframes can be from fisheye images, in which case the projection function is $\pi_{f}$, or from stereo images in which case the projection function is $\pi_{m}$ or $\pi_{s}$, depending on whether the keypoint is monocular or stereo, respectively.}

\subsection{Loop Closing}

For place recognition, we adapted DBoW2 to SIFT features and trained a million word vocabulary with ten branching factors and six levels, like the ORB vocabulary used in ORB-SLAM2. The vocabulary was trained on an extensive set of underwater imagery data, including the UWHandles and LizardIsland datasets presented in this paper, plus three large imagery datasets from the Australian Center for Field Robotics: Tasmania CSP~\cite{barrett2011methods}, Scott Reef 25~\cite{steinberg2010towards}, and Tasmania O'Hara 7~\cite{steinberg2011bayesian}. 2000 CudaSIFT features were extracted per image, with the images upscaled by a factor of 2 for the first scale pyramid level, the initial blur set to 1.6, and the difference of Guassian threshold set to 1.0.

After a loop closing event, a full bundle adjustment optimization is initiated, following the same optimization as Eq. (2), but including all keyframes and points in the map except for the origin stereo keyframe, which remains fixed.

\subsection{Datasets}

\subsubsection{Stereo Survey Dataset}

\begin{figure}
    \centering
    \begin{subfigure}[b]{0.4\columnwidth}
      \includegraphics[width=\linewidth]{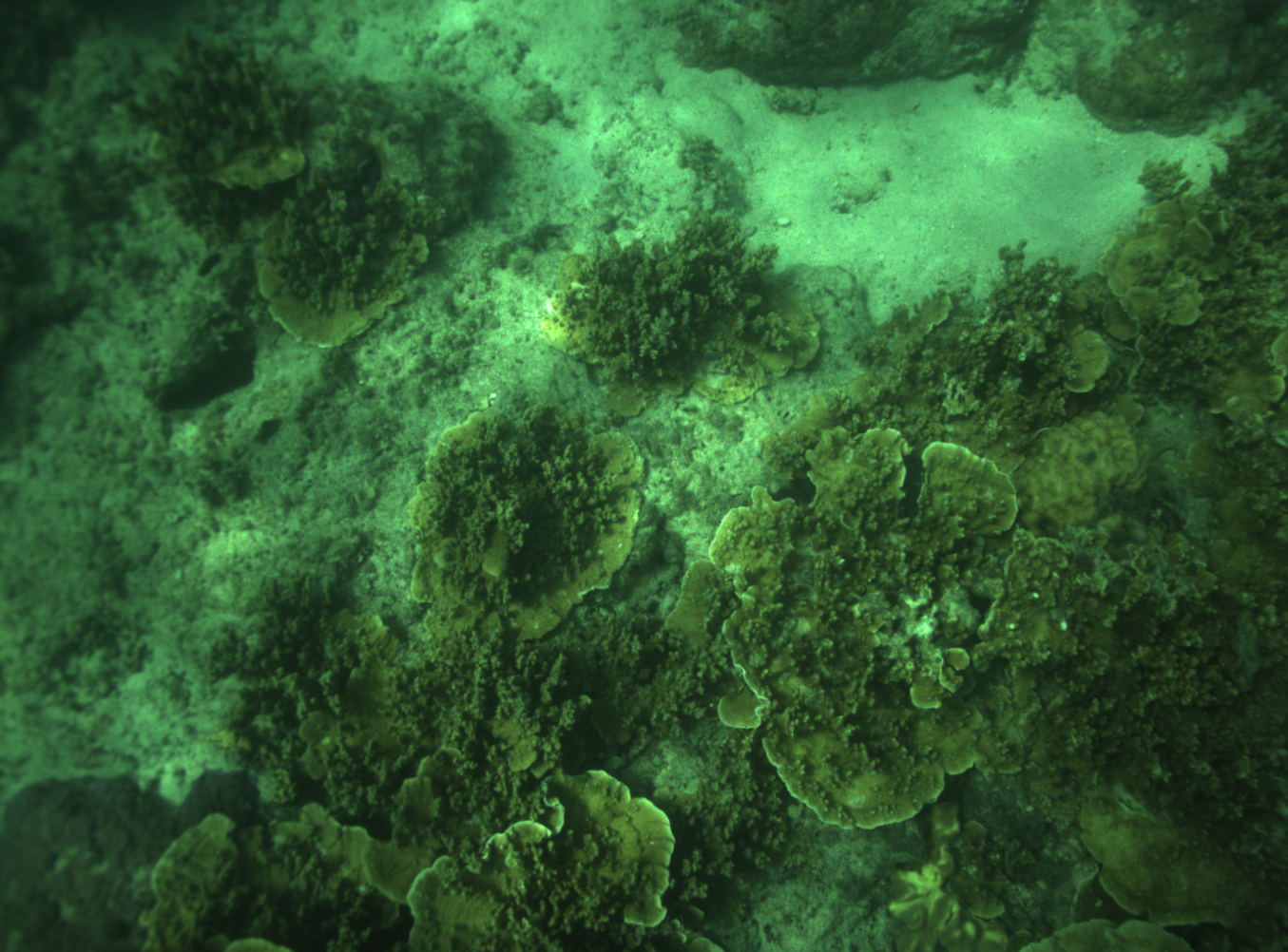}
      \caption{Left stereo image}
      \label{fig:spiral-image}
    \end{subfigure}
    \hfill 
    \begin{subfigure}[b]{0.55\columnwidth}
      \includegraphics[width=\linewidth]{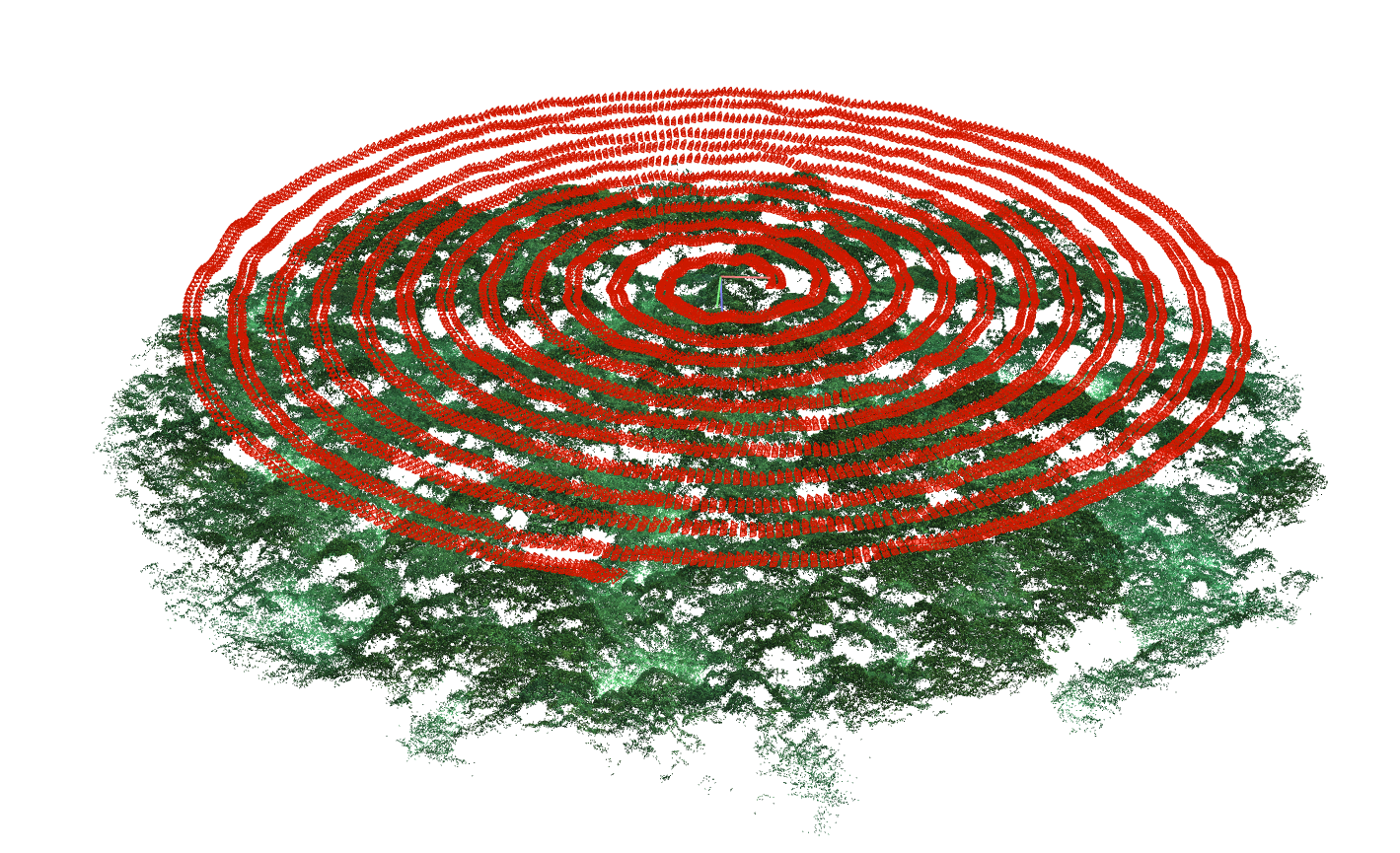}
      \caption{COLMAP reconstruction}
      \label{fig:spiral-colmap}
    \end{subfigure}

    \caption{The LizardIsland spiral survey dataset was collected with a diver operated stereo rig. The ground truth reconstruction was generated with COLMAP.}
    \label{fig:spiral-dataset}
    \vspace{-10pt}
\end{figure}

A stereo SLAM evaluation dataset was collected with a diver operated camera rig on a shallow coral reef of Lizard Island in Australia~(fig.~\ref{fig:spiral-dataset}). The dataset was collected using a spiral survey technique~\cite{pizarro2017simple} that fully covered a circular area of approximately 14m in diameter, with natural sunlight providing the only illumination. The rectified stereo image size is 1355x1002 pixels and the images were collected at 5Hz. We refer to this dataset as \textbf{LizardIsland}.

To obtain a ground truth comparison for evaluating our stereo SLAM method, we processed the dataset through COLMAP~\cite{schoenberger2016sfm} to generate a sparse 3D reconstruction with optimized camera poses. COLMAP does not fix the scale during optimization, so the reconstruction was scaled in post-process to match the mean left and right stereo pair baseline to the calibrated value.

\subsubsection{Hybrid Vehicle-Manipulator Dataset}

    


During a cruise in 2019, a hybrid dataset of synchronized vehicle mounted stereo and wrist mounted fisheye imagery was collected in natural deep ocean environments of the Costa Rican continental shelf margin with the SuBastian ROV, operated by Schmidt Ocean Institute~\cite{Vrolijk2121}. We have previously published the fisheye imagery portion of this data as the \textbf{UWHandles} dataset~\cite{billings2020silhonet}. For this work, we have extended this dataset by further processing four environmentally unique stereo and fisheye image sequences for evaluation of our hybrid SLAM method. We refer to these sequences as Mounds1, Mounds2, Seeps1, and Seeps2. For these sequences, TagSLAM~\cite{pfrommer2019tagslam} was used to obtain ground truth pose estimates for the stereo and fisheye cameras, based on the detection of AprilTags~\cite{wang2016iros} distributed in the scenes.

%% file: text/results.tex
All evaluations were run on a desktop computer with an AMD Ryzen Threadripper 2990WX CPU and an NVIDIA Titan V GPU.

\subsection{Comparative Feature Analysis}

We conducted an evaluation to determine which feature representation is best adapted to the visual degradation of underwater environments and can be robustly matched between hybrid perspective and fisheye frames with variable relative poses. We sampled every fifth hybrid frame from each of the UWHandles dataset sequences and, to reduce any bias from artificial features, we used the tracked AprilTag poses from TagSLAM to project circular masks over the tags in the image frames. For each feature type, 2000 features were extracted from each image, and the features were brute force matched across each hybrid fisheye and left stereo image pair. Lowe's ratio test was applied to remove ambiguous matches, with a ratio threshold of 0.8 for all feature types except ContextDesc, which achieved significantly improved performance with a ratio of 0.9. OpenCV's RANSAC based essential matrix fitting was used to filter the matches and recover a relative pose estimate between each fisheye and stereo frame. Table~\ref{tab:feature-eval} shows the results of this evaluation. Given that an essential matrix based pose estimate does not provide scale, both the orientation and translation errors of the pose estimates were evaluated as angular errors. For translation, this error is the angular difference between the translation direction vector from the left stereo frame to the fisheye frame. The performance was evaluated using the area under the accuracy-threshold curve (AUC) with a max angular error of 180\degree. While most of the tested feature types were popular conventional features, we also tested two deep learned feature variants: ContextDesc and SuperPoint. We note that the learned features were used with their provided model weights and were not fine-tuned on underwater data. Of the conventional feature types, ROOT\_SIFT and SIFT perform the best, achieving significantly better performance than ORB. Of the deep learned features, SuperPoint had highly variable performance across the different sequences, and the mean number of inlier matches was lower than other conventional features. Interestingly, ContextDesc performed the best overall out of all the feature types, consistently matching more than double the features of ROOT\_SIFT and achieving very high AUC scores. It is noteworthy that ContextDesc uses SIFT interest points but learns the descriptor, so all of the best performing features are based on the SIFT detector. These results merit further investigation into the application of learned features for underwater vision. For our initial implementation in this work, we chose to use a highly optimized GPU accelerated implementation of SIFT, but we note that the learned descriptors of ContextDesc are 128-d, like SIFT, and are directly compatible with our entire method pipeline.

\begin{table*}[!ht]
\centering
\scriptsize
\caption{Area under accuracy-threshold curve evaluation of feature matching performance on the UWHandles underwater hybrid image sequences. Accuracy is evaluated as angular error in the predicted rotation (AUC Rot) and translation direction vector (AUC Trans) between each hybrid left stereo and fisheye image pair. Also reported is the mean number of inlier feature matches across each sequence.}
\label{tab:feature-eval}
\begin{tabular}{llrrrrrrr}
\hline
Sequence & \multicolumn{1}{c}{} & \multicolumn{1}{l}{SIFT~\cite{lowe2004distinctive}} & \multicolumn{1}{l}{ROOT SIFT\cite{arandjelovic2012three}} & \multicolumn{1}{l}{ORB~\cite{rublee_orb_2011}} & \multicolumn{1}{l}{SURF~\cite{bay2008speeded}} & \multicolumn{1}{l}{AKAZE~\cite{alcantarilla2011fast}} & \multicolumn{1}{l}{CONTEXTDESC~\cite{luo2019contextdesc}} & \multicolumn{1}{l}{SUPERPOINT~\cite{detone2018superpoint}} \\ \hline
\multirow{3}{*}{Mounds1} & AUC Trans & 0.949 & 0.936 & 0.856 & 0.877 & 0.922 & 0.970 & \textbf{0.98} \\
 & AUC Rot & 0.937 & 0.948 & 0.750 & 0.812 & 0.858 & 0.951 & \textbf{0.964} \\
 & Mean Matches & 91 & 101 & 25 & 34 & 46 & \textbf{210} & 74 \\ \hline
\multirow{3}{*}{Mounds2} & AUC Trans & 0.946 & 0.940 & 0.864 & 0.886 & 0.906 & \textbf{0.976} & 0.947 \\
 & AUC Rot & 0.810 & 0.853 & 0.488 & 0.676 & 0.629 & \textbf{0.959} & 0.873 \\
 & Mean Matches & 31 & 33 & 14 & 21 & 21 & \textbf{80} & 41 \\ \hline
\multirow{3}{*}{Seeps1} & AUC Trans & 0.964 & 0.980 & 0.885 & 0.869 & 0.904 & \textbf{0.986} & 0.938 \\
 & AUC Rot & 0.944 & 0.965 & 0.745 & 0.770 & 0.811 & \textbf{0.980} & 0.885 \\
 & Mean Matches & 64 & 73 & 23 & 28 & 43 & \textbf{150} & 53 \\ \hline
\multirow{3}{*}{Seeps2} & AUC Trans & 0.960 & 0.964 & 0.935 & 0.930 & 0.954 & \textbf{0.974} & 0.917 \\
 & AUC Rot & 0.942 & 0.953 & 0.894 & 0.891 & 0.926 & \textbf{0.965} & 0.763 \\
 & Mean Matches & 89 & 100 & 60 & 50 & 92 & \textbf{146} & 39 \\ \hline
\end{tabular}
\vspace{-10pt}
\end{table*}

\begin{figure}
    \centering
    \begin{subfigure}[t]{0.49\columnwidth}
      \includegraphics[width=\linewidth]{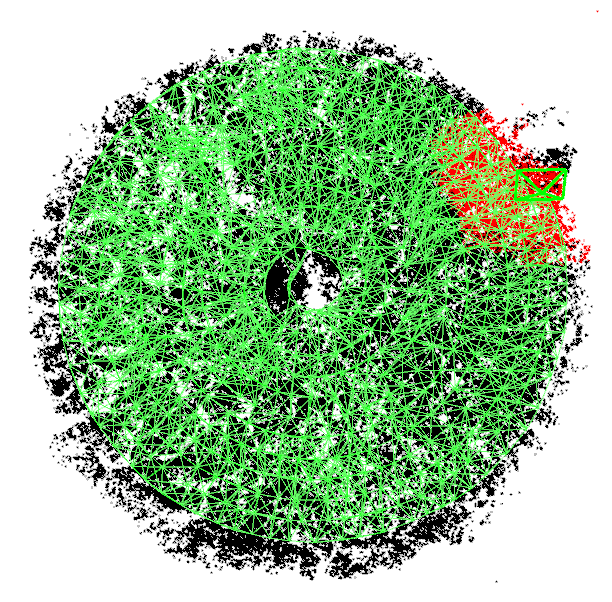}
      \caption{4k features/frame}
      \label{fig:spiral-map-4000}
    \end{subfigure}
    \hfill 
    \begin{subfigure}[t]{0.49\columnwidth}
      \includegraphics[width=\linewidth,trim={0.7cm 0.7cm 0.7cm 0.7cm},clip]{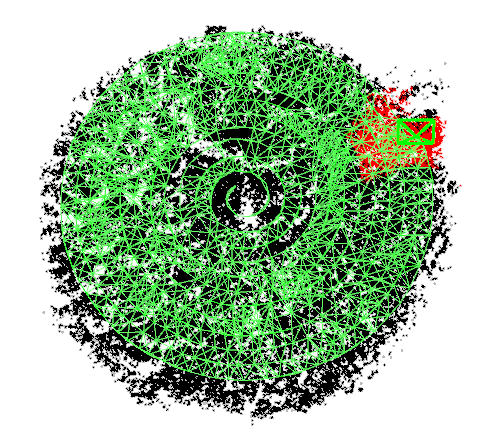}
      \caption{2k features/frame}
      \label{fig:spiral-map-2000}
    \end{subfigure}
    \caption{Final stereo SLAM maps on the LizardIsland dataset, showing the densely connected keyframe graphs.}
    \label{fig:spiral-map}
    \vspace{-8pt}
\end{figure}

\subsection{Stereo SLAM}

\begin{figure}
    \centering
    \begin{subfigure}[t]{0.49\columnwidth}
      \includegraphics[width=\linewidth,trim={0 4.5cm 0 3.5cm},clip]{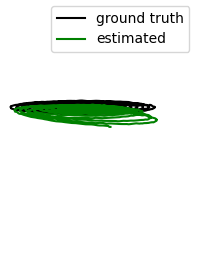}
      \caption{Tracking}
      \label{fig:spiral-track-side}
    \end{subfigure}
    \hfill 
    \begin{subfigure}[t]{0.49\columnwidth}
      \includegraphics[width=\linewidth,trim={0 4.5cm 0 3.5cm},clip]{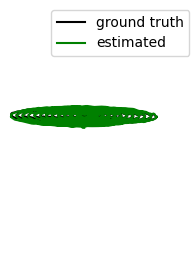}
      \caption{SLAM}
      \label{fig:spiral-slam-side}
    \end{subfigure}

    \caption{Stereo SLAM results (green) for the LizardIsland dataset when loop closing is disabled~(\subref{fig:spiral-track-side}) and enabled~(\subref{fig:spiral-slam-side}), plotted against the ground truth (black).}
    \label{fig:spiral-results}
    \vspace{-15pt}
\end{figure}

The core of our system is a stereo SLAM pipeline, which must be robust to underwater environments. We used the LizardIsland survey dataset to evaluate the stereo SLAM performance. We tested ORB-SLAM2 on this dataset, both with and without loop closing enabled, but it lost track after only a few frames and was unable to relocalize. We also tested the vanilla VISO2 stereo odometer, but, even with extensive tuning, VISO2 failed to track the dataset with sensible accuracy. \revision{As a result, the stereo SLAM method of \cite{negre_cluster-based_2016}, specifically designed for underwater environments, also failed on our dataset, as it runs on top of VISO2.} We evaluated our stereo SLAM method with both 2000 and 4000 CudaSIFT features extracted each frame, with an interest point Difference of Gaussian threshold of 1.2. Figure~\ref{fig:spiral-results} shows the results for 4000 features, both with and without loop closing enabled, and table~\ref{tab:spiral-results} gives the performance of the system in all tests. The test trajectories were aligned with the COLMAP ground truth using the Horn method~\cite{horn1987closed} without scaling. The figure shows that our visual odometer based tracking method without loop closing tracks very well in the horizontal plane with most of the drift error being accumulated in the z-depth estimate. For both extracted feature counts, the table shows that a high accuracy, with less than 2cm root mean squared absolute trajectory error (RMSE), is attained by the full SLAM system with loop closing. For 4000 features per frame, the number of map points in the final map is approximately double the map point count for 2000 features per frame, showing that the system scales well with the number of extracted features. The system can achieve \textgreater10Hz for 2000 features per frame, which is a high framerate for underwater systems. Figure~\ref{fig:spiral-map} shows the densely connected keyframe graphs for the final SLAM maps, demonstrating consistent loop closing between neighboring spiral trajectories.

\begin{table}[]
\centering
\caption{Stereo SLAM performance on the LizardIsland dataset, with the number of extracted features is set to 4000 and 2000. Performance is evaluated as RMSE of the absolute trajectory error. Results are reported with and without loop closing enabled. Also reported is the number of keyframes (KFs) and map points (MPs) in the final map and the average frame processing time in the tracking thread.}
\label{tab:spiral-results}
\setlength\tabcolsep{4pt}
\begin{tabular}{lcccc}
\hline
System Mode & RMSE (cm) & KFs & MPs & \multicolumn{1}{l}{Avg Time (ms)} \\ \hline
Tracking Only (4000) & 49.1 & - & - & 94.2 \\
Loop Closing (4000) & 1.4 & 562 & 190,474 & 117.9 \\
Tracking Only (2000) & 58.2 & - & - & 55.7 \\
Loop Closing (2000) & 1.8 & 622 & 98,812 & 64.8 \\ \hline
\end{tabular}
\vspace{0pt}
\end{table}

\subsection{Hybrid SLAM}

\begin{table}[h]
\centering
\caption{Hybrid SLAM timing evaluation, measured as the mean frame processing time in the tracking thread.}
\label{tab:hybrid-timing}
\begin{tabular}{l|ccc}
\hline
\# Features / Frame & 2k & 4k & 6k \\
Mean Time & 179ms & 249ms & 314ms
\end{tabular}
\vspace{-10pt}
\end{table}

\begin{figure*}[h]
    \centering
    \begin{subfigure}[t]{0.24\linewidth}
      \includegraphics[width=\linewidth]{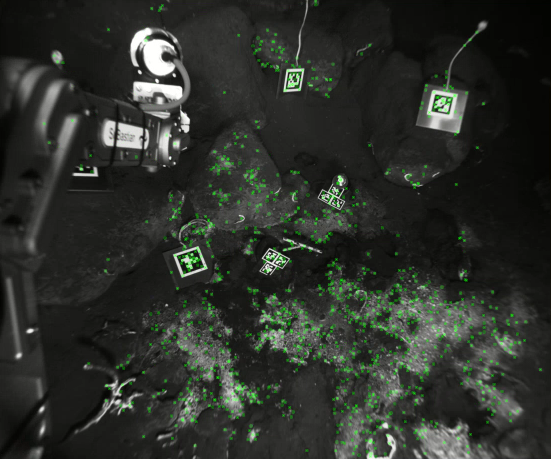}
    \end{subfigure}
    \begin{subfigure}[t]{0.24\linewidth}
      \includegraphics[width=\linewidth]{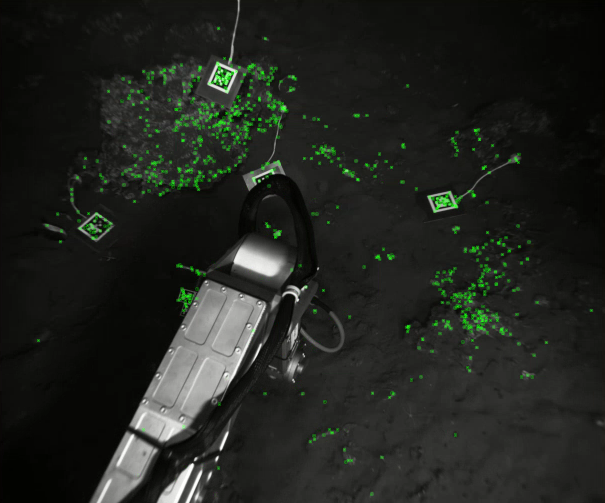}
    \end{subfigure}
    \begin{subfigure}[t]{0.24\linewidth}
      \includegraphics[width=\linewidth]{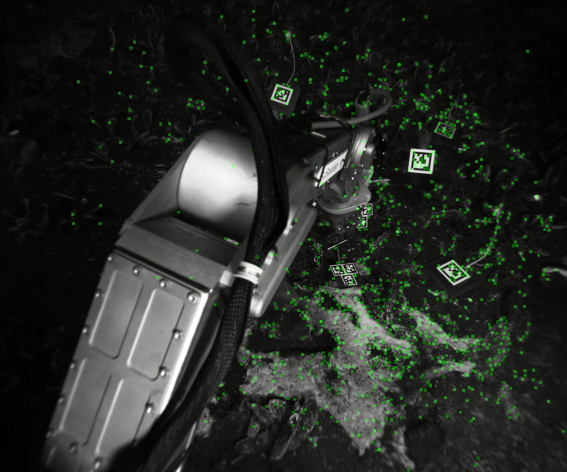}
    \end{subfigure}
    \begin{subfigure}[t]{0.24\linewidth}
      \includegraphics[width=\linewidth]{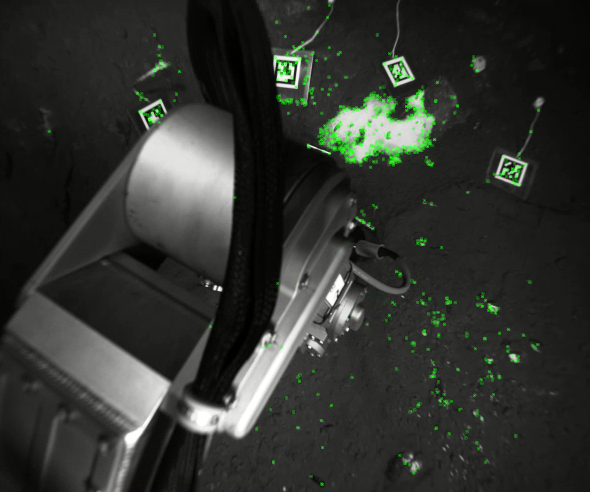}
    \end{subfigure}
    
    \begin{subfigure}[t]{0.24\linewidth}
      \includegraphics[width=\linewidth]{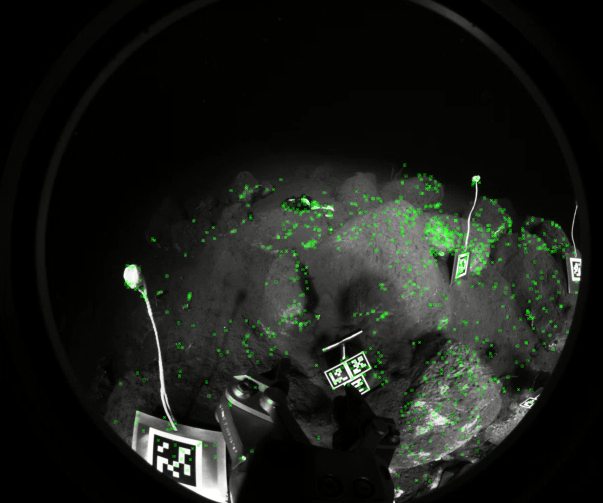}
    \end{subfigure}
    \begin{subfigure}[t]{0.24\linewidth}
      \includegraphics[width=\linewidth]{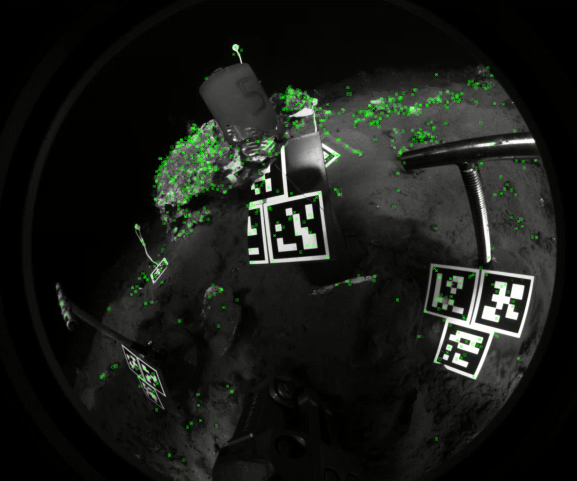}
    \end{subfigure}
    \begin{subfigure}[t]{0.24\linewidth}
      \includegraphics[width=\linewidth]{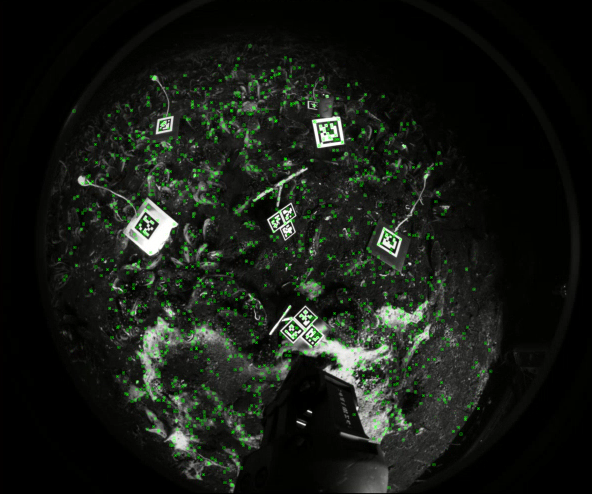}
    \end{subfigure}
    \begin{subfigure}[t]{0.24\linewidth}
      \includegraphics[width=\linewidth]{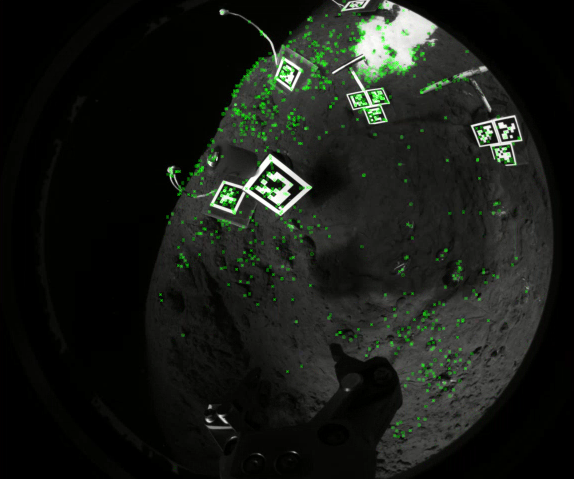}
    \end{subfigure}
    
    \begin{subfigure}[t]{0.24\linewidth}
      \includegraphics[width=\linewidth]{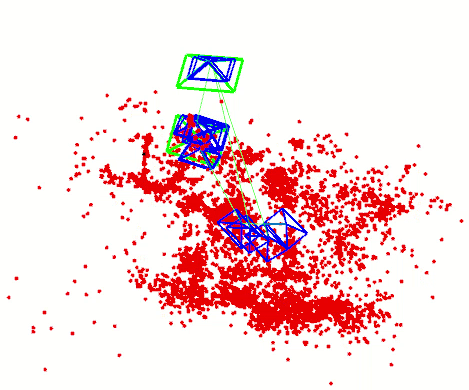}
      \caption{Mounds1}
    \end{subfigure}
    \begin{subfigure}[t]{0.24\linewidth}
      \includegraphics[width=\linewidth]{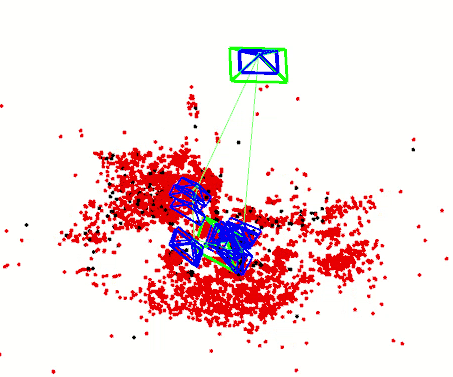}
      \caption{Mounds2}
    \end{subfigure}
    \begin{subfigure}[t]{0.24\linewidth}
      \includegraphics[width=\linewidth]{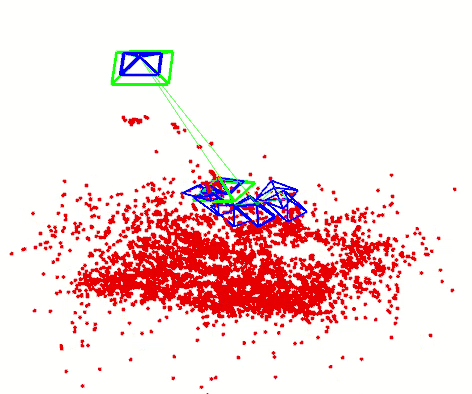}
      \caption{Seeps1}
    \end{subfigure}
    \begin{subfigure}[t]{0.24\linewidth}
      \includegraphics[width=\linewidth]{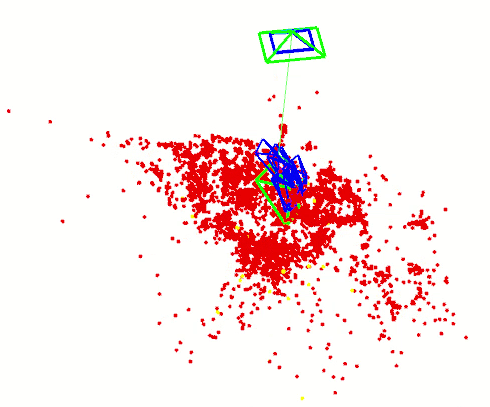}
      \caption{Seeps2}
    \end{subfigure}

    \caption{Snapshots of hybrid SLAM running on the UWHandles sequences. Top row is the left stereo camera frame, middle row is the manipulator mounted fisheye frame, and bottom row is the map with the keypoints and keyframes.}
    \label{fig:uwhandles-snaps}
    \vspace{-5pt}
\end{figure*}

We evaluated the performance of our hybrid SLAM system on the four sequences of the UWHandles dataset. The results are reported in table~\ref{tab:hybrid-results}. For all sequences, every stereo frame was successfully registered in the SLAM map. Given that the stereo camera is mostly stationary across these image sequences, and to reduce the effect of noise in the imperfect ground truth, we evaluated the hybrid SLAM error using the relative pose estimates between the left stereo and fisheye cameras for each synchronized hybrid frame. Only hybrid frames where the fisheye frame was successfully registered into the map were included in the error evaluation. As is detailed in the table, the system generated approximately twice as many keyframes and map points when running in hybrid mode versus stereo only mode, demonstrating the ability to extend the map beyond the limited stereo camera viewpoint. Also, the majority of fisheye frames were successfully registered into the map for all sequences. Despite the sequences varying significantly in environment type, the hybrid SLAM mode is able to generate a similar amount of keyframes and map points for each sequence, and the estimated pose errors are very similar across each sequence, demonstrating our system can operate in challenging and diverse, natural seafloor environments. Figure~\ref{fig:uwhandles-snaps} shows a frame capture from running hybrid SLAM on each of the sequences. Table~\ref{tab:hybrid-timing} provides the timing evaluation for processing a hybrid stereo and fisheye frame pair through the tracking thread for different feature count settings. For 4000 features extracted per image, the system can easily attain 3Hz, which is the rate that the UWHandles data was collected.

\begin{table*}[!ht]
\centering
\caption{Evaluation of hybrid SLAM on the UWHandles dataset. Error is evaluated on the estimated pose difference between the left stereo and fisheye cameras for each synchronized hybrid frame, where $\Delta t$ is translation error and $\Delta q$ is rotation error. The "hybrid matches" column gives the number of fisheye frames registered in the map over the total number of frames in the sequence. The error is only evaluated over the registered frames. The "KFs" column is the number of keyframes in the final map for the hybrid SLAM mode versus stereo only mode, and the "MPs" column is the same format for the number of final keypoints in the map.}
\label{tab:hybrid-results}
\begin{tabular}{lccccccc}
\hline
Sequence & $\Delta t$ mean (cm) & $\Delta t$ median (cm) & $\Delta q$ mean (deg) & \multicolumn{1}{l}{$\Delta q$ median (deg)} & hybrid matches & KFs hybrid/stereo & MPs hybrid/stereo \\ \hline
Mounds1 & 2.04 & 2.06 & 0.58 & 0.50 & 652 / 783 & 21 / 11 & 4271 / 2671 \\
Mounds2 & 1.38 & 1.22 & 0.98 & 0.82 & 713 / 756 & 24 / 11 & 4086 / 1773 \\
Seeps1 & 2.82 & 2.02 & 1.17 & 0.46 & 1059 / 1089 & 24 / 11 & 4636 / 2847 \\
Seeps2 & 2.12 & 1.84 & 1.38 & 0.97 & 778 / 802 & 23 / 16 & 4320 / 2365 \\ \hline
\end{tabular}
\vspace{-10pt}
\end{table*}

%% file: text/conclusion.tex
We have presented a novel hybrid SLAM method, targeting deployment on underwater vehicle manipulator systems, that can operate in real-time. The method can fuse features from both a vehicle mounted stereo camera and a manipulator mounted fisheye camera into the same map, enabling dynamic viewpoint acquisition and map extension with the manipulator mounted camera. We have demonstrated the robustness of our method on both a shallow reef stereo image survey dataset and on four hybrid image sequences captured in natural, deep seafloor environments.
\revision{In this work, the system has been tested on a desktop computer, which is suitable for ROV operations, where the compute can run topside through a tethered connection with the vehicle. In future work, we will optimize the method to run on embedded GPU devices which can be deployed onboard the vehicle. Currently, the system fuses the fisheye data into the map without consideration of the manipulator state.} In future work, we will also explore the formulation of a kinematic factor from the manipulator joint states between the fisheye and the stereo camera to improve registration of the fisheye camera into the map and the overall robustness of the mapping method. This factor would also enable real-time feedback for the kinematic calibration of the manipulator, which is a challenging problem for the imprecise hydraulic manipulators common for underwater systems. We will also explore the use of learned feature descriptors such as ContextDesc to improve system performance. Finally, we will explore methods for generating dense reconstructions based on the sparse feature maps and optimized camera poses to build a complete real-time scene reconstruction method for UVMSs.